\pdfoutput=1

\documentclass[11pt]{article}

\usepackage[final]{acl}

\usepackage{times}
\usepackage{latexsym}

\usepackage[T1]{fontenc}

\usepackage[utf8]{inputenc}

\usepackage{microtype}

\usepackage{inconsolata}

\usepackage{graphicx}

\usepackage{graphicx} 
\usepackage{multirow}
\usepackage{makecell}
\usepackage{enumitem}
\usepackage{amsmath}
\usepackage{algorithm}
\usepackage{algpseudocode}
\usepackage{booktabs}
\usepackage{amsmath}
\usepackage{multirow}
\usepackage{bbding}
\usepackage{mathrsfs}
\usepackage{subfigure}
\usepackage{amssymb}
\usepackage{mathtools}
\usepackage{xcolor}
\usepackage{colortbl}
\usepackage{color}
\usepackage[most]{tcolorbox}
\usepackage{hyperref}
\usepackage{listings}
\usepackage[most]{tcolorbox}

\definecolor{deepgreen}{RGB}{0, 70, 0}
\definecolor{deepred}{RGB}{182, 32, 22}

\title{ProgCo: Program Helps Self-Correction of Large Language Models}

\author{Xiaoshuai Song\thanks{Work done during internship at Alibaba Inc.}, Yanan Wu\thanks{Corresponding Author: Yanan Wu}, Weixun Wang, Jiaheng Liu, Wenbo Su, Bo Zheng\\
  Taobao \& Tmall Group of Alibaba\\
  \texttt{\{songxiaoshuai.sxs,lixing.wyn\}@alibaba-inc.com}
}

\begin{document}
\maketitle
\begin{abstract}
Self-Correction aims to enable large language models  (LLMs) to self-verify and self-refine their initial responses without external feedback.
However, LLMs often fail to effectively self-verify and generate correct feedback, further misleading refinement and leading to the failure of self-correction, especially in complex reasoning tasks.
In this paper, we propose \textbf{Prog}ram-driven Self-\textbf{Co}rrection (\textbf{ProgCo}). 
First, \textbf{prog}ram-driven \textbf{ve}rification (\textbf{ProgVe}) achieves complex verification logic and extensive validation through self-generated, self-executing verification pseudo-programs.
Then, \textbf{prog}ram-driven \textbf{re}finement (\textbf{ProgRe}) receives feedback from ProgVe, conducts dual reflection and refinement on both responses and verification programs to mitigate misleading of incorrect feedback in complex reasoning tasks.
Experiments on three instruction-following and mathematical benchmarks indicate that ProgCo achieves effective self-correction, and can be further enhance performance when combined with real program tools. 
We release our code at \url{https://github.com/songxiaoshuai/progco}.
\end{abstract}

\section{Introduction}
\label{sec:intro}

\begin{figure}[t]
    \centering
    \resizebox{0.48\textwidth}{!}{
    \includegraphics{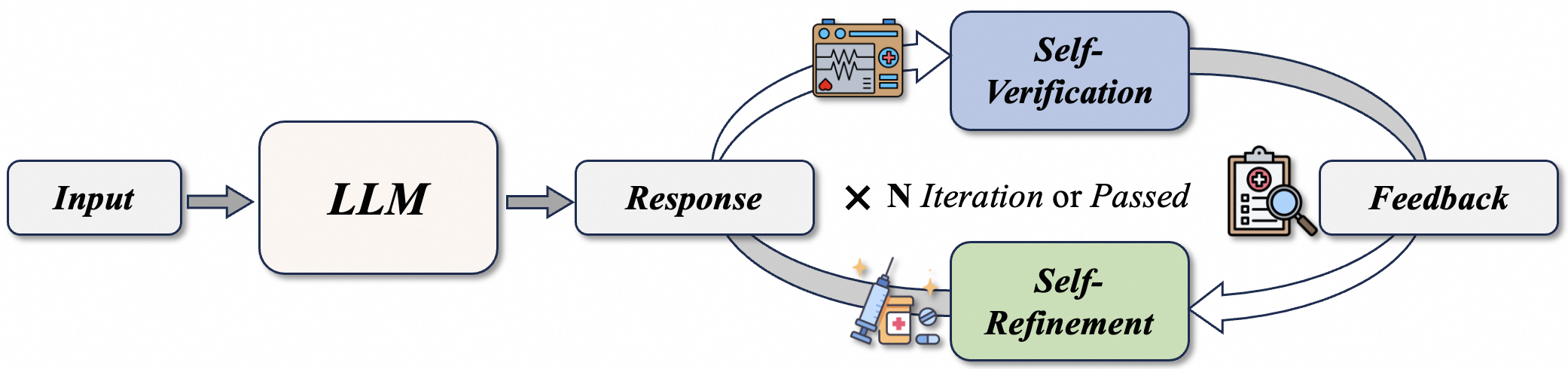}}
    \caption{Illustration of a typical workflow of LLM’s intrinsic self-correction.}
    \label{fig:task_intro}
\end{figure}

Although large language models (LLMs) have shown excellent performance on certain tasks, they still face challenges such as hallucinations and unfaithful reasoning when solving complex instruction-following and reasoning tasks~\citep{zhao2023survey, chang2024survey}. 
Self-correction is an expected capability of LLMs, wherein the LLM first needs to reflect on its initial output, identify potential issues and generate feedback (\textit{self-verification} phase), which then guides the LLM to optimize and refine its output (\textit{self-refinement} phase), as illustrated in Fig \ref{fig:task_intro}~\citep{pan-etal-2024-automatically,kumar2024traininglanguagemodelsselfcorrect}. 
However, studies have shown that current LLMs severely lack this capability and struggle to achieve effective self-correction in the absence of external feedback (also called \textit{intrinsic self-correction}), particularly in complex tasks~\citep{li-etal-2024-hindsight,huang2024large,tyen-etal-2024-llms,kamoi-etal-2024-llms}.

A core reason for the failure of self-correction in LLMs is their inability to effectively self-detect problematic outputs and generate high-quality feedback~\citep{tyen-etal-2024-llms}. 
Existing works mainly used two approaches for self-verification: (1) prompting LLMs to perform step-by-step self-checks~\citep{madaan2024self}, or (2) generating a checklist based on the task and then checking responses against this list~\citep{zhang-etal-2024-self-contrast,cook2024tickingboxesgeneratedchecklists}. 
However, on one hand, LLMs often exhibit overconfidence, making it difficult for them to identify their own errors or hallucinations, resulting in a low recall of incorrect responses~\citep{zhang-etal-2024-self-contrast}. On the other hand, for complex tasks, these methods struggle to parse intricate verification logic. For instance, checklists usually only express parallel relationships and examine superficial issues. Ineffective verification further leads to ineffective refinement. The inaccurate and low-quality error detection and feedback not only makes it challenging for LLMs to correct erroneous outputs in one attempt but also seriously mislead them into modifying from correct to incorrect for misrecalled responses.
To overcome this issue, we propose \textbf{Prog}ram-driven Self-\textbf{Co}rrection (\textbf{ProgCo}), achieving effective self-correction by incorporating self-generated and self-executed programs in the verification and refinement stages.

\begin{figure}[t]
    \centering
    \resizebox{0.48\textwidth}{!}{
    \includegraphics{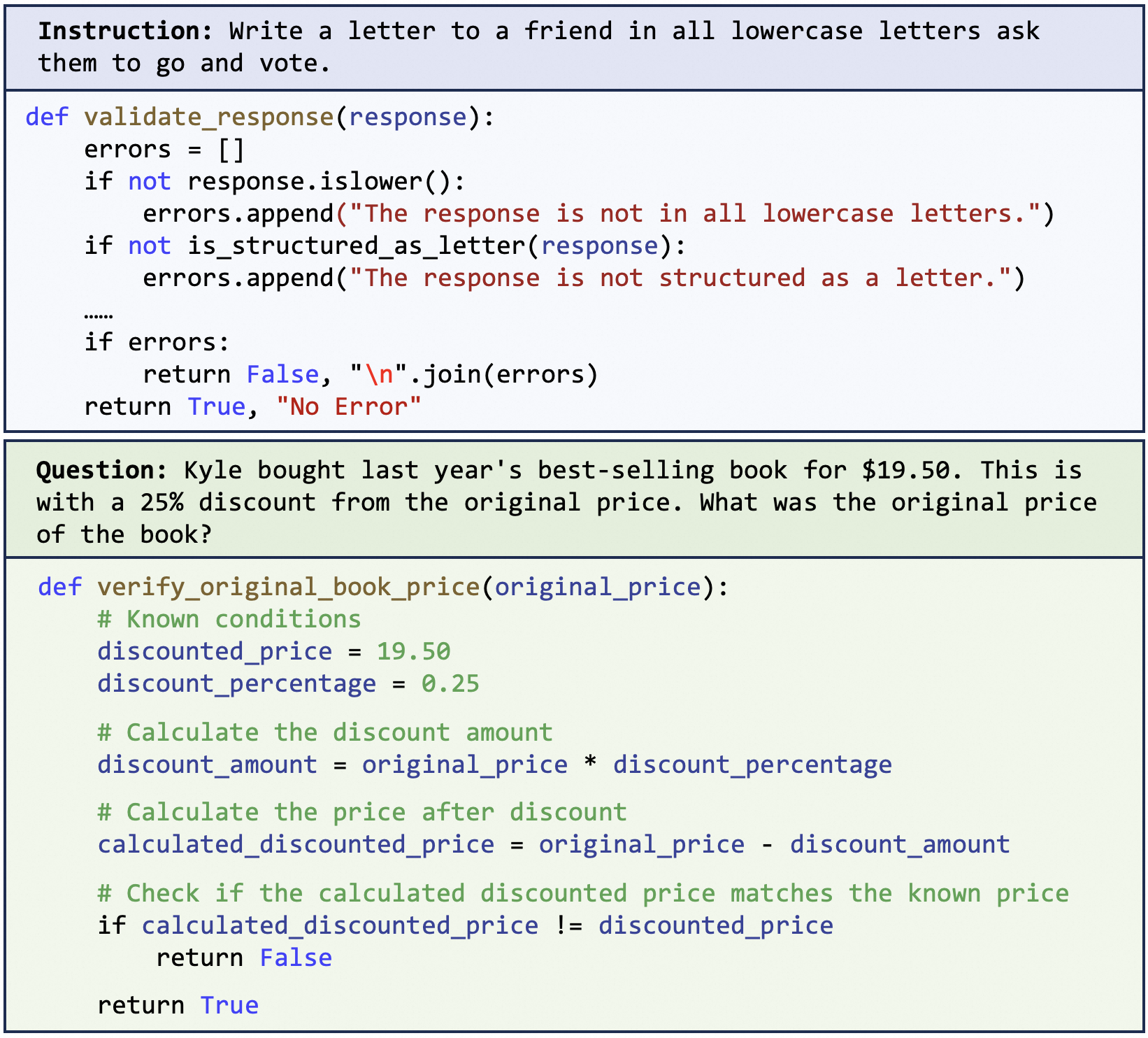}}
    \caption{Illustration of generating verification pseudo-program for input tasks.}
    \label{fig:code_verfication_case}
\end{figure}

We first introduce \textbf{Prog}ram-driven \textbf{Ve}rification (\textbf{ProgVe}) to achieve better self-verification. Different from studies like PAL and POT~\citep{pmlr-v202-gao23f,chen2023program,gou2024tora}, which integrate program or symbolic solvers for forward reasoning, ProgVe focuses on the self and reverse verification phase. 
In general, the LLM is initially prompted to generate a pseudo verification program for the input task. After obtaining the initial response, the LLM further acts as a program-executor, executing the verification program step by step to obtain the verification results.
The motivation for ProgVe stems from: (1) Compared to the ambiguity of natural language, code program can express more complex verification logic and structures. (2) Using LLM as a program-executor not only allows a focus on the verification logic of the code without requiring strict executability, but also incorporate LLM's own knowledge and causal understanding into execution, such as the virtual function  \texttt{is\_structured\_as\_letter} in Fig \ref{fig:code_verfication_case}.

To address misleading self-refinement caused by incorrect self-verification in complex reasoning tasks such as mathematics, we further introduce \textbf{Prog}ram-driven \textbf{Re}finement (\textbf{ProgRe}), featuring a dual refinement mechanism for both response and program.
Within the framework, to avoid directly misleading of feedback, the response revised based on feedback is treated as an intermediate and contrasted with the pre-revision response to identify differences and generate insights, which will help in regenerating the final refined response.
To more fundamentally address incorrect verification, besides refining response with grogram, the verification program is also reflected upon and optimized with the assistance of response information.

We provide a detailed exposition of proposed method in Section \ref{sec:method}. In Section \ref{sec:experiment}, we demonstrate that our method can achieve effective self-correction and outperforms all baselines in correcting instruction-following and mathematical reasoning tasks.  A series of analyses further provides insights into self-correction and our method from multiple perspectives. 

In summary, we propose ProgCo for effective self-correction,  consisting of two components: ProgVe and ProgRe. Our contributions are three-fold:
(1) We propose ProgVe, a method enabling LLMs to self-generate and self-execute validation programs for self-verification.
(2)We propose ProgRe, a self-refinement method that is robust to incorrect feedback and combines dual optimization of response and verification programs.
(3) Experiments and analyses on three datasets verify the effectiveness of our method in self-correction.

\section{Method}
\label{sec:method}

\subsection{Program-driven Verification}
\label{subsec:ProgVe}

Given model $M$ and input $x$, in the $i$-th iteration of self-correction, the goal of self-verification is to generate feedback $fb_i$ for response $y_i$, indicating whether $y_i$ passes verification or the reason it fails.

\textbf{Verification Program Generation.}
After obtaining the initial response $y_0 = M(x)$, we first use prompt $P^{gen}_{fuc}$ to guide $M$ in generating a verification pseudo-program function $f$. This process is independent of $y_0$ to ensure a different perspective and avoid biases from the response:
\begin{equation}
f=M(P^{gen}_{fuc}||x)
\end{equation}
As illustrated in Fig \ref{fig:code_verfication_case}, for instruction-following task, $f$ verifies a series of constraints extracted from $x$. For mathematical problems, $f$ starts from the output answer and uses reverse reasoning to verify step-by-step whether it contradicts the given conditions in $x$.

\textbf{Verification Program Execution.}
For each round, we use prompt $P^{exec}_{fuc}$ to instruct $M$ act as a code executor, taking $y_{i}$ as input, executing $f$ step by step to obtain the execution result $r_i$. $r_i$ is further converted to feedback $fb_{i}$ by prompt $P_{fb}$:
\begin{equation}
r_{i}=M(P^{exec}_{fuc}||x||y_i), fb_{i}=M(P_{fb}||x||y_i||r_i)
\end{equation}
If the verification passes, self-correction stops and yields the final output $y_{final} = y_i$. Otherwise, if $fb_{i}$ indicates that $y_i$ fails to meet constraints or contains contradictions, the process enters the self-refinement stage until maximum rounds $I$ are reached.

\begin{figure}[t]
    \centering
    \resizebox{0.48\textwidth}{!}{
    \includegraphics{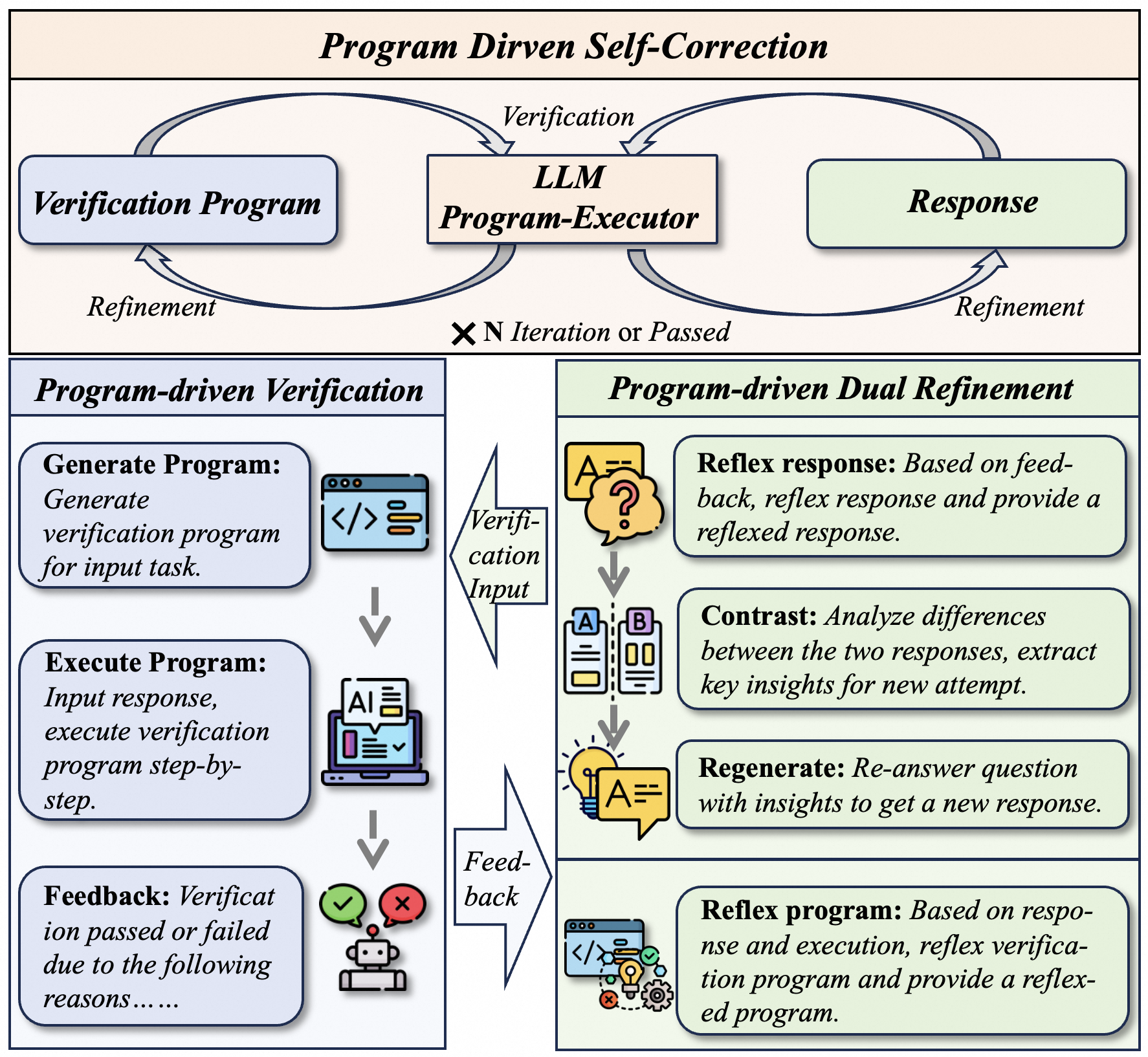}}
    \caption{The overall framework of ProgCo, achieving self-correction through iterative ProgVe and ProgRe.}
    \label{fig:refine_framework}
\end{figure}

\subsection{Program-driven Refinement}
\label{subsec:prore}
The \textbf{vanilla self-refinement} generates a new output $y_{i+1}=M(p_{refine}||x||y_{i}||fb_{i})$ through prompt $p_{refine}$. This method is effective for tasks like instruction following, as constraints are apparent, and the correct program $f$ and clear feedback $fb_{i}$ are easily obtained. However, in reasoning tasks such as mathematics, the correct $f$ and $fb_{i}$ are difficult to obtain in one attempt and will mislead refinement, as described in Section \ref{sec:intro}.
Therefore, we introduce the Program-driven Refinement framework to address this, as shown in Fig \ref{fig:refine_framework}.

\textbf{Preliminary Reflection.}
Unlike directly refining $y_i$ using $fb_{i}$, we prompt $M$ to first reflect on $y_i$ under the feedback and then output a temporary response $y^{temp}_{i+1}$, which can either maintain $y_i$ unchanged or update it:
\begin{equation}
y^{temp}_{i+1}=M(P_{reflex}||x||y_i||fb_i)
\end{equation}

\textbf{Contrast and Regenerate.}
If a new answer is obtained, $y^{temp}_{i+1}$ is further compared with $y_i$ to identify differences. These differences will be transformed into insights $ins$ for problem-solving, assisting in regenerating the refined response:\\
\begin{equation}
ins=M(P_{cont}||y_{i}||y^{temp}_{i+1}) , y_{i+1}=M(ins||x)
\end{equation} 

\textbf{Verification Program Refinement.}
To correct potentially incorrect validation code generated in the initial round, we further introduce self-refinement of the validation code. Utilizing information from $y_i$ and $fb_i$, the validation code $f_i$ will self-reflect and generate new validation code $f_{i+1}$ for next round of self-verification:
\begin{equation}
f_{i+1}=M(P^{code}_{reflex}||x||f_{i}||y_{i}||fb_{i})
\end{equation} 

We provide detailed pseudo-code and prompts used in the proposed method in Appendix \ref{app:method_details}.

\begin{table*}[!t]
\scriptsize 
    \centering
\setlength{\tabcolsep}{1.2mm}{
    \begin{tabular}{lcccccccccccc}
        \toprule
        \multirow{2}{*}{\textbf{Method}} & \multicolumn{4}{c}{\textbf{Llama3.1-8B-Instruct}} & \multicolumn{4}{c}{\textbf{GPT-3.5}} & \multicolumn{4}{c}{\textbf{GPT-4o}} \\
        \cmidrule(r){2-5}\cmidrule(lr){6-9}\cmidrule(lr){10-13}
         & IF (Pr) & IF (Ins) & GSM8K & MATH & IF (Pr) & IF (Ins) & GSM8K & MATH & IF (Pr) & IF (Ins) & GSM8K & MATH \\
        \midrule
        Initial Score & 73.75 & 81.29 & 85.82 & 46.8 & 58.23 & 68.35 & 76.5 & 36.2 & 82.99 & 87.52 & 95.07 & 77.2 \\
        \midrule
        \multicolumn{13}{c}{\textit{Maximum One Round of Self-Correction}}\\
        \midrule
Vanilla-reflex & 
$\text{68.39}_{\textcolor{deepred}{\text{-5.36}}}$ & 
$\text{76.74}_{\textcolor{deepred}{\text{-4.55}}}$ & 
$\text{83.32}_{\textcolor{deepred}{\text{-2.50}}}$ & 
$\text{44.6}_{\textcolor{deepred}{\text{-2.20}}}$ & 
$\text{54.9}_{\textcolor{deepred}{\text{-3.33}}}$ & 
$\text{69.18}_{\textcolor{deepgreen}{\text{+0.83}}}$ & 
$\text{72.4}_{\textcolor{deepred}{\text{-4.10}}}$ & 
$\text{35.6}_{\textcolor{deepred}{\text{-0.60}}}$ & 
$\text{83.73}_{\textcolor{deepgreen}{\text{+0.74}}}$ & 
$\text{88.37}_{\textcolor{deepgreen}{\text{+0.85}}}$ & 
$\text{93.31}_{\textcolor{deepred}{\text{-1.76}}}$ & 
$\text{77.6}_{\textcolor{deepgreen}{\text{+0.40}}}$ \\
Self-Refine & 
$\text{75.79}_{\textcolor{deepgreen}{\text{+2.04}}}$ & 
$\text{82.49}_{\textcolor{deepgreen}{\text{+1.20}}}$ & 
$\text{84.69}_{\textcolor{deepred}{\text{-1.13}}}$ & 
$\text{40.6}_{\textcolor{deepred}{\text{-6.20}}}$ & 
$\text{59.33}_{\textcolor{deepgreen}{\text{+1.10}}}$ & 
$\text{68.94}_{\textcolor{deepgreen}{\text{+0.59}}}$ & 
$\text{76.35}_{\textcolor{deepred}{\text{-0.15}}}$ & 
$\text{38.0}_{\textcolor{deepgreen}{\text{+1.80}}}$ & 
$\text{84.66}_{\textcolor{deepgreen}{\text{+1.67}}}$ & 
$\text{88.97}_{\textcolor{deepgreen}{\text{+1.45}}}$ & 
$\text{94.92}_{\textcolor{deepred}{\text{-0.15}}}$ & 
$\text{75.4}_{\textcolor{deepred}{\text{-1.80}}}$ \\
Self-Reflection & 
$\text{73.38}_{\textcolor{deepred}{\text{-0.37}}}$ & 
$\text{80.22}_{\textcolor{deepred}{\text{-1.07}}}$ & 
$\text{84.23}_{\textcolor{deepred}{\text{-1.59}}}$ & 
$\text{43.8}_{\textcolor{deepred}{\text{-3.00}}}$ & 
$\text{59.7}_{\textcolor{deepgreen}{\text{+1.47}}}$ & 
$\text{68.94}_{\textcolor{deepgreen}{\text{+0.59}}}$ & 
$\text{71.95}_{\textcolor{deepred}{\text{-4.55}}}$ & 
$\text{34.8}_{\textcolor{deepred}{\text{-1.40}}}$ & 
$\text{84.84}_{\textcolor{deepgreen}{\text{+1.85}}}$ & 
$\text{88.97}_{\textcolor{deepgreen}{\text{+1.45}}}$ & 
$\text{95.0}_{\textcolor{deepred}{\text{-0.07}}}$ & 
$\text{77.6}_{\textcolor{deepgreen}{\text{+0.40}}}$ \\
CheckList & 
$\text{74.49}_{\textcolor{deepgreen}{\text{+0.74}}}$ & 
$\text{81.89}_{\textcolor{deepgreen}{\text{+0.60}}}$ & 
$\text{84.76}_{\textcolor{deepred}{\text{-1.06}}}$ & 
$\text{47.0}_{\textcolor{deepgreen}{\text{+0.20}}}$ & 
$\text{59.15}_{\textcolor{deepgreen}{\text{+0.92}}}$ & 
$\text{69.06}_{\textcolor{deepgreen}{\text{+0.71}}}$ & 
$\text{77.63}_{\textcolor{deepgreen}{\text{+1.13}}}$ & 
$\text{36.0}_{\textcolor{deepred}{\text{-0.20}}}$ & 
$\text{85.4}_{\textcolor{deepgreen}{\text{+2.41}}}$ & 
$\text{89.45}_{\textcolor{deepgreen}{\text{+1.93}}}$ & 
$\text{94.84}_{\textcolor{deepred}{\text{-0.23}}}$ & 
$\text{77.4}_{\textcolor{deepgreen}{\text{+0.20}}}$ \\
\textbf{ProgCo (Ours)} & 
\textbf{$\text{76.34}_{\textcolor{deepgreen}{\text{+2.59}}}$} & 
\textbf{$\text{83.69}_{\textcolor{deepgreen}{\text{+2.40}}}$} & 
\textbf{$\text{86.58}_{\textcolor{deepgreen}{\text{+0.76}}}$} & 
\textbf{$\text{50.2}_{\textcolor{deepgreen}{\text{+3.40}}}$} & 
\textbf{$\text{62.85}_{\textcolor{deepgreen}{\text{+4.62}}}$} & 
\textbf{$\text{71.58}_{\textcolor{deepgreen}{\text{+3.23}}}$} & 
\textbf{$\text{82.34}_{\textcolor{deepgreen}{\text{+5.84}}}$} & 
\textbf{$\text{42.0}_{\textcolor{deepgreen}{\text{+5.80}}}$} & 
\textbf{$\text{87.99}_{\textcolor{deepgreen}{\text{+5.00}}}$} & 
\textbf{$\text{91.85}_{\textcolor{deepgreen}{\text{+4.33}}}$} & 
\textbf{$\text{95.75}_{\textcolor{deepgreen}{\text{+0.68}}}$} & 
\textbf{$\text{79.4}_{\textcolor{deepgreen}{\text{+2.20}}}$} \\
        \midrule
        \multicolumn{13}{c}{\textit{Maximum Three Rounds of Self-Correction}}\\
        \midrule
Vanilla-reflex & 
$\text{65.25}_{\textcolor{deepred}{\text{-8.50}}}$ & 
$\text{74.82}_{\textcolor{deepred}{\text{-6.47}}}$ & 
$\text{81.43}_{\textcolor{deepred}{\text{-4.39}}}$ & 
$\text{44.6}_{\textcolor{deepred}{\text{-2.20}}}$ & 
$\text{55.27}_{\textcolor{deepred}{\text{-2.96}}}$ & 
$\text{65.83}_{\textcolor{deepred}{\text{-2.52}}}$ & 
$\text{71.8}_{\textcolor{deepred}{\text{-4.70}}}$ & 
$\text{35.8}_{\textcolor{deepred}{\text{-0.40}}}$ & 
$\text{84.29}_{\textcolor{deepgreen}{\text{+1.30}}}$ & 
$\text{88.73}_{\textcolor{deepgreen}{\text{+1.21}}}$ & 
$\text{94.08}_{\textcolor{deepred}{\text{-0.99}}}$ & 
$\text{79.2}_{\textcolor{deepgreen}{\text{+2.00}}}$ \\
Self-Refine & 
$\text{76.34}_{\textcolor{deepgreen}{\text{+2.59}}}$ & 
$\text{82.73}_{\textcolor{deepgreen}{\text{+1.44}}}$ & 
$\text{84.99}_{\textcolor{deepred}{\text{-0.83}}}$ & 
$\text{43.4}_{\textcolor{deepred}{\text{-3.40}}}$ & 
$\text{59.7}_{\textcolor{deepgreen}{\text{+1.47}}}$ & 
$\text{69.06}_{\textcolor{deepgreen}{\text{+0.71}}}$ & 
$\text{76.35}_{\textcolor{deepred}{\text{-0.15}}}$ & 
$\text{38.0}_{\textcolor{deepgreen}{\text{+1.80}}}$ & 
$\text{85.03}_{\textcolor{deepgreen}{\text{+2.04}}}$ & 
$\text{89.45}_{\textcolor{deepgreen}{\text{+1.93}}}$ & 
$\text{94.92}_{\textcolor{deepred}{\text{-0.15}}}$ & 
$\text{76.4}_{\textcolor{deepred}{\text{-0.80}}}$ \\
Self-Reflection & 
$\text{73.75}_{\textcolor{deepgreen}{\text{+0.00}}}$ & 
$\text{81.06}_{\textcolor{deepred}{\text{-0.23}}}$ & 
$\text{84.53}_{\textcolor{deepred}{\text{-1.29}}}$ & 
$\text{48.0}_{\textcolor{deepgreen}{\text{+1.20}}}$ & 
$\text{59.52}_{\textcolor{deepgreen}{\text{+1.29}}}$ & 
$\text{69.18}_{\textcolor{deepgreen}{\text{+0.83}}}$ & 
$\text{74.07}_{\textcolor{deepred}{\text{-2.43}}}$ & 
$\text{35.6}_{\textcolor{deepred}{\text{-0.60}}}$ & 
$\text{85.77}_{\textcolor{deepgreen}{\text{+2.78}}}$ & 
$\text{89.69}_{\textcolor{deepgreen}{\text{+2.17}}}$ & 
$\text{94.92}_{\textcolor{deepred}{\text{-0.15}}}$ & 
$\text{78.4}_{\textcolor{deepgreen}{\text{+1.20}}}$ \\
CheckList & 
$\text{75.05}_{\textcolor{deepgreen}{\text{+1.30}}}$ & 
$\text{82.61}_{\textcolor{deepgreen}{\text{+1.32}}}$ & 
$\text{84.91}_{\textcolor{deepred}{\text{-0.91}}}$ & 
$\text{47.4}_{\textcolor{deepgreen}{\text{+0.60}}}$ & 
$\text{59.52}_{\textcolor{deepgreen}{\text{+1.29}}}$ & 
$\text{69.3}_{\textcolor{deepgreen}{\text{+0.95}}}$ & 
$\text{77.79}_{\textcolor{deepgreen}{\text{+1.29}}}$ & 
$\text{35.8}_{\textcolor{deepred}{\text{-0.40}}}$ & 
$\text{85.77}_{\textcolor{deepgreen}{\text{+2.78}}}$ & 
$\text{89.69}_{\textcolor{deepgreen}{\text{+2.17}}}$ & 
$\text{95.0}_{\textcolor{deepred}{\text{-0.07}}}$ & 
$\text{77.0}_{\textcolor{deepred}{\text{-0.20}}}$ \\
\textbf{ProgCo (Ours)} & 
\textbf{$\text{77.82}_{\textcolor{deepgreen}{\text{+4.07}}}$} & 
\textbf{$\text{84.29}_{\textcolor{deepgreen}{\text{+3.00}}}$} & 
\textbf{$\text{87.41}_{\textcolor{deepgreen}{\text{+1.59}}}$} & 
\textbf{$\text{50.6}_{\textcolor{deepgreen}{\text{+3.80}}}$} & 
\textbf{$\text{63.03}_{\textcolor{deepgreen}{\text{+4.80}}}$} & 
\textbf{$\text{71.82}_{\textcolor{deepgreen}{\text{+3.47}}}$} & 
\textbf{$\text{83.78}_{\textcolor{deepgreen}{\text{+7.28}}}$} & 
\textbf{$\text{44.2}_{\textcolor{deepgreen}{\text{+8.00}}}$} & 
\textbf{$\text{87.8}_{\textcolor{deepgreen}{\text{+4.81}}}$} & 
\textbf{$\text{91.97}_{\textcolor{deepgreen}{\text{+4.45}}}$} & 
\textbf{$\text{95.75}_{\textcolor{deepgreen}{\text{+0.68}}}$} & 
\textbf{$\text{80.0}_{\textcolor{deepgreen}{\text{+2.80}}}$} \\
        \bottomrule
    \end{tabular}}
    \caption{The result of different self-correction methods. The metric for GSM8K and MATH is accuracy, IF (Pr) and IF (Ins) denote IFEval's strict prompt and instruction metrics, respectively. $\textcolor{deepgreen}{\text{+}}$/$\textcolor{deepred}{\text{-}}$ indicates change from initial score.}
    \label{tab:main_results}
\end{table*}

\section{Experiment}
\label{sec:experiment}

\subsection{Experiment Setup}
We evaluate ProgCo on the instruction-following dataset IFEval~\citep{zhou2023instructionfollowingevaluationlargelanguage} and the mathematics datasets GSM8K~\citep{cobbe2021trainingverifierssolvemath} and MATH~\citep{hendrycksmath2021}\footnote{We randomly sample 500 instances from MATH test set.}.
Since complex reasoning is not involved, we only use the combination of ProgVe and vanilla refinement for IFEval.
Additionally, we calculate the score for each self-correction round individually, without assuming the next round is correct just because the previous one was. This may differ from some evaluation settings.
We provide detailed datasets, baseline introductions, and implementation details in Appendix \ref{app:experiment_details}.

\subsection{Main Result}
Table \ref{tab:main_results} shows the performance comparison between ProgCo and different self-correction baselines. 
\footnote{Due to varied focuses, we discuss various inference methods in Appendix \ref{app:related_work} and compare the self-consistency sampling methods in Section  \ref{subapp:self_consistency_experiment}.}
Overall, ProgCo outperforms all baselines with a large margin across three benchmarks. On GPT-3.5, ProgCo improves over the initial response by 4.62\% (IFEval(Pr)), 3.23\% (IFEval(Ins)), 5.84\% (GSM8K), and 5.8\% (MATH) with just one round of self-correction. After three rounds, the improvements further increase to 4.80\% (IFEval(Pr)), 3.47\% (IFEval(Pr)),7.28\% (GSM8K) and 8.0\% (MATH). Similar improvements are observed on other LLMs.
From a task perspective, many baselines achieve positive improvements on IFEval but failed on mathematical tasks, while our method achieve significant positive improvements on both GSM8K and MATH, demonstrating its effectiveness in complex reasoning.

\subsection{Ablation Analysis}

\begin{figure}[t]
    \centering
    \resizebox{0.48\textwidth}{!}{
    \includegraphics{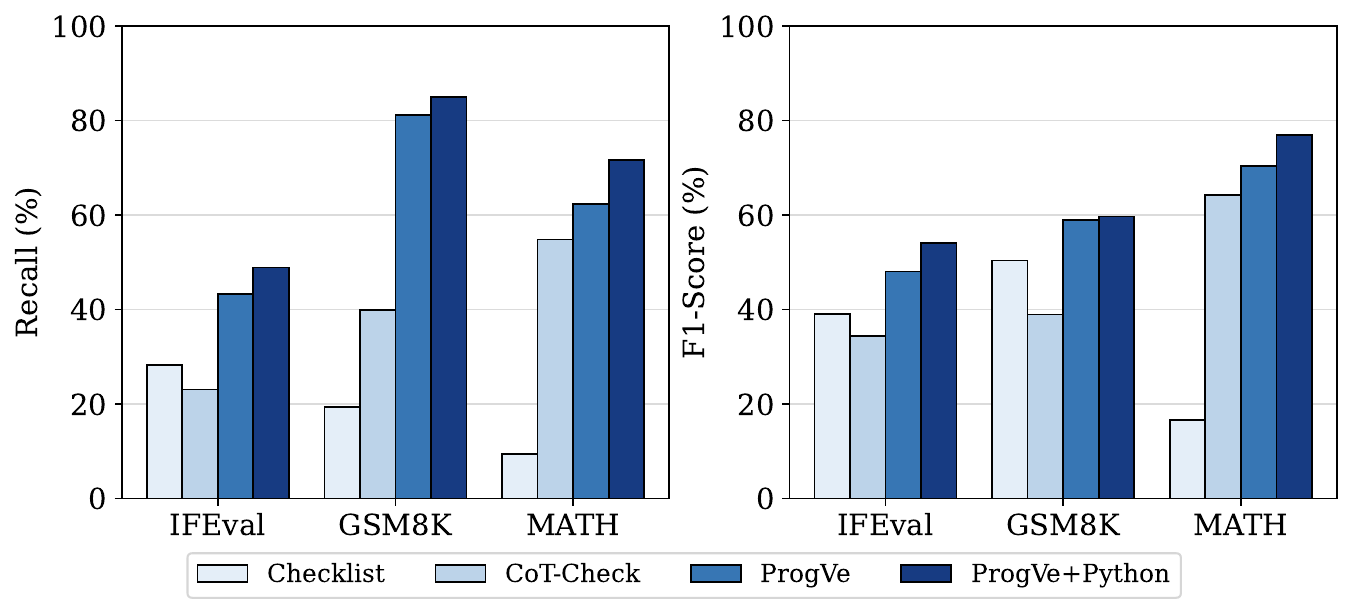}}
    \caption{Recall and F1 scores of self-verification methods for incorrect responses on GPT-3.5.}
    \label{fig:recall}
\end{figure}

\textbf{Recall of Self-Verification.}
Fig \ref{fig:recall} shows the recall and F1-score for CoT-Check, Checklist, and ProgVe in identifying incorrect responses. ProgVe outperforms the baselines in both Recall and F1-score significantly, with further improvements when combined with Python tool. This demonstrates the advantages of using self-play programs in verification, including expressing more complex structures, providing a different perspective, and integrate with symbolic tools.

\begin{table}[t]
\scriptsize
    \centering
\setlength{\tabcolsep}{0.8mm}{
    \begin{tabular}{lcccc}
        \toprule
        \textbf{Method} & \textbf{ACC} $\boldsymbol{\uparrow}$ & \textbf{Avg Turn} $\boldsymbol{\downarrow}$ & $\boldsymbol{\Delta^{i \to c}} \boldsymbol{\uparrow}$ &  $\boldsymbol{\Delta^{c \to i}} \boldsymbol{\downarrow}$ \\
        \midrule
        ProgVe & 83.78 & 0.88 & 47.75 & 15.42 \\
        \quad \textit{w/o} cont. \& regen. & 
        $\text{79.08}_{\textcolor{deepred}{\text{-4.7}}}$ & 
        $\text{0.87}_{\textcolor{deepgreen}{\text{-0.01}}}$ & 
        $\text{43.88}_{\textcolor{deepred}{\text{-3.87}}}$ & 
        $\text{35.18}_{\textcolor{deepred}{\text{+19.76}}}$ \\
        \quad \textit{w/o} program reflex & 
        $\text{83.02}_{\textcolor{deepred}{\text{-0.76}}}$ & 
        $\text{1.06}_{\textcolor{deepred}{\text{+0.18}}}$ & 
        $\text{44.98}_{\textcolor{deepred}{\text{-2.77}}}$ & 
        $\text{15.95}_{\textcolor{deepred}{\text{+0.53}}}$ \\
    \quad \textit{w/o} feedback & 
    $\text{81.35}_{\textcolor{deepred}{\text{-2.43}}}$ & 
    $\text{0.90}_{\textcolor{deepred}{\text{+0.02}}}$ & 
    $\text{30.5}_{\textcolor{deepred}{\text{-17.25}}}$ & 
    $\text{7.36}_{\textcolor{deepgreen}{\text{-8.06}}}$ \\
        \bottomrule
    \end{tabular}
}
    \caption{Ablation on GPT-3.5 with maximum three rounds of self-correction. Avg Turn is the average rounds of self-refinement for all samples. $\Delta^{i \to c}$ is the ratio of incorrect-to-correct transitions among recalled ground-truth incorrect samples, and $\Delta^{c \to i}$ is the opposite for recalled correct samples. Avg Turn denotes the average ProgVe turns for all samples.}
    \label{tab:ablation}
\end{table}

\begin{table}[t]
\scriptsize
    \centering
\setlength{\tabcolsep}{2.2mm}{
    \begin{tabular}{lcccc}
        \toprule
        \multirow{2}{*}{\textbf{Model}} & \multicolumn{2}{c}{\textbf{IFEval (Pr})} & \multicolumn{2}{c}{\textbf{MATH (ACC)}} \\ 
        \cmidrule(lr){2-3}\cmidrule(lr){4-5}
         & ProgVe & ProgVe+Python & ProgVe & ProgVe+Python \\ \midrule
        GPT-3.5 & 62.0 & \textbf{$\text{64.14}_{\textcolor{deepgreen}{\text{+2.14}}}$} & 44.2 & \textbf{$\text{44.6}_{\textcolor{deepgreen}{\text{+0.4}}}$} \\ 
        GPT-4o  & 87.8 & \textbf{$\text{91.31}_{\textcolor{deepgreen}{\text{+3.51}}}$} & 80.0 & \textbf{$\text{81.2}_{\textcolor{deepgreen}{\text{+1.2}}}$}\\ \bottomrule
    \end{tabular}
}
    \caption{Performance  of ProgCo in integrating the Python executor tool during the ProgVe process.}
    \label{tab:python_tool}
\end{table}

\textbf{Ablation on Self-Refinement.}
We conduct an ablation analysis of ProgRe in Table \ref{tab:ablation}.
First, removing contrast and regeneration leads to a significant increase in the ratio of correction to incorrection, proving that this strategy can effectively alleviate feedback misleading.
Second, since wrong verification programs continually trap responses from passing verification, removing the reflection of program leads to a significant increase in the average rounds of self-correction.
Lastly, without program feedback involved in response reflection, the ratios of both correct and incorrect refinement notably decrease, indicating issues of LLM’s overconfidence and misleading feedback, respectively.

\subsection{Integrating symbolic tools for LLM program-executor.} 
Due to the advantage of program easily integrating with symbolic tools, we further indicate in Prompt \(P^{exec}_{fuc}\) that LLM executor can delegate complex numerical operations to an actual python tool to overcome the shortcomings of LLM.
As shown in Table \ref{tab:python_tool}, this further improves ProgCo's performance.
For example, ProgVe obtains more precise feedback on constraints such as word count and keywords, thereby significantly improving the performance of IFEval.

\subsection{Effect of Self-Correction Iterations.}
\begin{figure}[t]
    \centering
    \resizebox{0.48\textwidth}{!}{
    \includegraphics{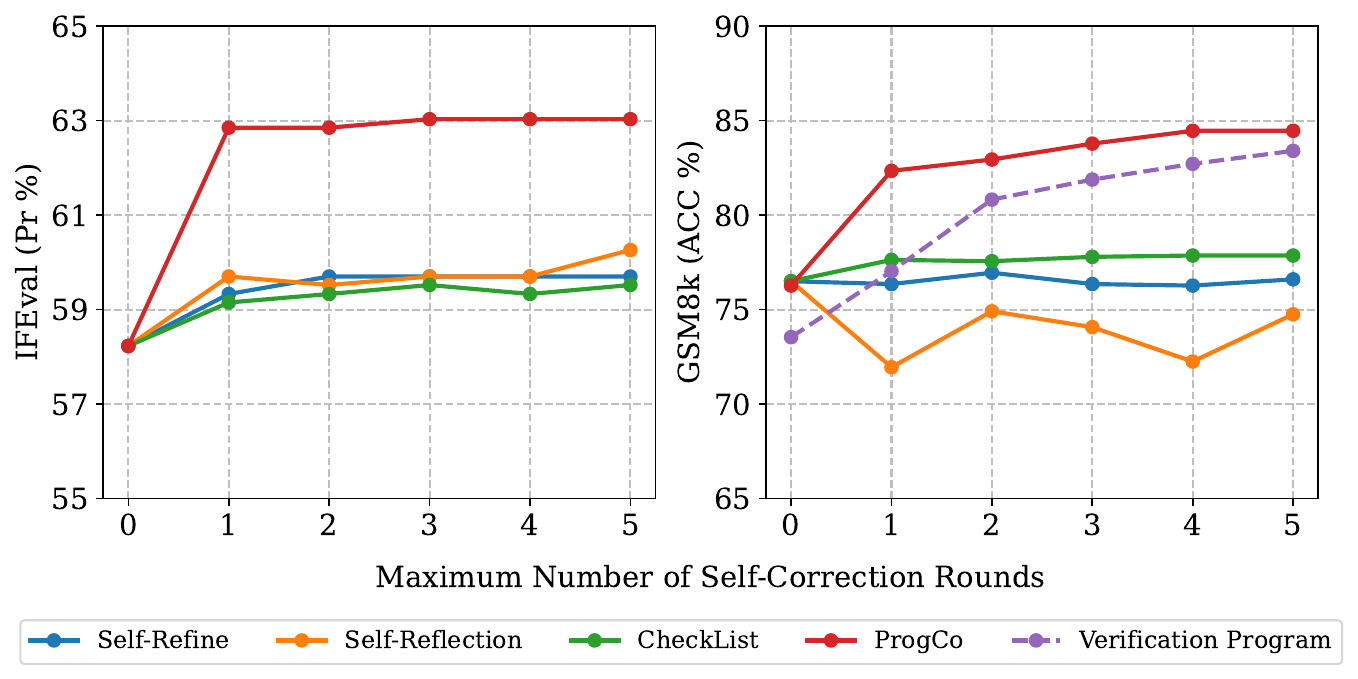}}
    \caption{Score variation with the maximum number of self-correction rounds on GPT-3.5. The program's ACC measures the consistency accuracy between the execution results of ProgVe and ground-truth scores.}
    \label{fig:turn}
\end{figure}

As shown in Fig \ref{fig:turn}, with increasing self-correction rounds,  baselines' performance either slightly improves or fluctuates near the initial score. In contrast, ProgCo shows a consistent upward trend, especially in the first few rounds. The continual improvement in the verification program's consistency accuracy further highlights the advantages of dual refinement.

\subsection{Comparison with Self-Consistency}
\label{subapp:self_consistency_experiment}
We choose the self-consistency (SC) as the sampling approach. For multiple sampled responses, SC-Vote determines the final answer based on majority voting, SC-Reflex reflects to arrive at a final response, and SC-Select prompts the LLM to compare and select one as the final response.
The results are shown in Table \ref{tab:sc}. On IFEval, SC-Vote is not applicable, and both SC-Reflex and SC-Select lead to poorer performance, indicating that sampling is not suitable for tasks like instruction-following. In contrast, ProgCo significantly improves instruction-following performance.
On mathematical tasks where sampling excels, ProgCo with one correction round outperforms 3-sampling significantly and 5-sampling on MATH. ProgCo with five correction rounds matches 10-sampling on GSM8K and slightly surpasses it on MATH.
It should be emphasized that, due to early stopping from ProgVe, ProgRe only performs an average of about 1.2 rounds of self-refinement (for instance, many cases pass self-verification without entering the self-refinement phase).

\begin{table}[t]
\scriptsize
    \centering
\setlength{\tabcolsep}{1.2mm}{
    \begin{tabular}{lcccc}
    \toprule
    \textbf{Method} & \textbf{MAX Turn} & \textbf{IFEval(Pr)} & \textbf{GSM8K} & \textbf{MATH} \\
    \midrule
    Initial Score & -- & 58.23 & 76.5 & 36.2 \\
    \midrule
    \multirow{3}{*}{SC-Vote} & 3 & -- & 79.45 & 37.4 \\
     & 5 & -- & 83.02 & 40.8 \\
     & 10 & -- & \textbf{84.46} & 44.2 \\
     \midrule
    \multirow{3}{*}{SC-Reflex} & 3 & 41.96 & 79.61 & 38.2 \\
     & 5 & 42.88 & 81.73 & 40.4 \\
     & 10 & 44.36 & 84.08 & 41.6 \\
     \midrule
     \multirow{3}{*}{SC-Select} & 3 & 55.82 & 77.41 & 36.2 \\
     & 5 & 55.64 & 78.24 & 38.6 \\
     & 10 & 56.19 & 78.09 & 36.0 \\
     \midrule
    \multirow{5}{*}{Ours} & 1 & \makecell{62.85\\(0.34 $\text{turn}_\text{avg}$)} & \makecell{82.34\\(0.41 $\text{turn}_\text{avg}$)} & \makecell{42.0\\(0.49 $\text{turn}_\text{avg}$)} \\
    & 3 & \makecell{63.03\\(0.80 $\text{turn}_\text{avg}$)} & \makecell{83.78\\(0.88 $\text{turn}_\text{avg}$)} & \makecell{44.2\\(0.97 $\text{turn}_\text{avg}$)} \\
    & 5 & \makecell{\textbf{63.03}\\(1.18 $\text{turn}_\text{avg}$)} & \makecell{\textbf{84.46}\\(1.13 $\text{turn}_\text{avg}$)} & \makecell{\textbf{44.4}\\(1.34 $\text{turn}_\text{avg}$)} \\
    \bottomrule
\end{tabular}
}
    \caption{Comparison of ProgCo and the Self-Consistency (SC) on GPT-3.5. The temperature for SC is 0.7. $\text{Turn}_\text{avg}$ denotes the average self-refinement (ProgRe) iterations for all samples.}
    \label{tab:sc}
\end{table}

\subsection{Case Study}
\label{subapp:case_study}
We present examples of ProgVe performing self-verification on IFEval and GSM8K in Figures \ref{fig:ifeval_case} and \ref{fig:gsm8k_case_verification}, respectively. Figure \ref{fig:gsm8k_case_refine} shows an example of PogRe performing self-refinement on GSM8K.
In Fig \ref{fig:ifeval_case}, the verification function not only generates executable programs such as \texttt{response!=response.upper()}, but also generates more vaguely defined programs like \texttt{is\_english(response)} and \texttt{has\_title(response)}. However, the LLM is able to successfully execute these pseudo-programs, surpassing the capabilities of a real program executor.
In Fig \ref{fig:gsm8k_case_verification}, the verification program verifies the correctness of the response in reverse. It starts with the predicted \texttt{remaining\_speed}, calculates the \texttt{average\_speed} step by step, and checks whether it matches the given condition \texttt{target\_average\_speed}.
In Fig \ref{fig:gsm8k_case_refine}, although the response still contains errors upon initial reflection, the LLM ultimately provides the correct answer to the question with the help of contrasted insights. Meanwhile, the verification program is also further optimized during the refinement process.

\section{Conclusion}
In this paper, we propose ProgCo, a program-driven self-correction method.
ProgCo first self-generates and self-executes verification pseudo-programs for self-verification (ProgVe), then uses dual reflection and refinement of responses and programs for self-refinement (ProgRe).
Experiments and analyses on three benchmarks demonstrate the effectiveness of ProgCo in self-correction.

\section*{Limitations}
In this paper, we propose ProgCo and experimentally validate its effective self-correction. However, there are still some limitations as follows:
(1) In terms of application scenarios, although using pseudo-program and LLM executors can extend the application scope beyond numerical and symbolic solving tasks, we primarily validated the effectiveness of ProgCo in instruction-following and mathematical tasks.
(2) One advantage of using LLMs in executing verification programs is the integration of their own knowledge and causal logic. However, they are limited in large and precise numerical calculations. This issue can be mitigated by combining real symbolic tools, as shown in the experiment in Table \ref{tab:python_tool}.
(3) Due to the lack of specialized training, we use detailed prompts to guide the LLM in completing tasks in ProgCo, which results in additional inference costs. Synthesizing data for each component of ProgCo and jointly training the LLM can replace the need for prompts and demonstration costs during inference.

\bibliography{custom}

\appendix


\section{Related Work}
\label{app:related_work}

\textbf{Self-Correction.}
Self-correction aims to enable LLMs to achieve the ability to self-check and correct its outputs~\citep{pan-etal-2024-automatically}.
Although some work~\citep{shinn2024reflexion,renze2024self} achieves correction by relying on environmental feedback (such as True/False signals), many studies~\citep{li-etal-2024-hindsight,huang2024large,tyen-etal-2024-llms,kamoi-etal-2024-llms} have shown that in the complete absence of environmental feedback, LLMs find it difficult to engage in effective self-reflection, with a particular emphasis on the challenge of identifying their own errors. To this end, some work~\citep{han2024smalllanguagemodelselfcorrect,zhang2024learningcheckunleashingpotentials,kumar2024traininglanguagemodelsselfcorrect} focuses on enhancing LLMs' self-correction capabilities during the training phase through imitation learning or reinforcement learning, while another part~\citep{madaan2024self,dhuliawala-etal-2024-chain,wu-etal-2024-large,zhang-etal-2024-self-contrast,kim2024language} focuses on designing reflection or correction pipelines during the inference phase to help LLMs examine and analyze their own outputs.
In this work, we focus on the inference phase, and to the best of our knowledge, we are the first to introduce self-generated and self-executed pseudo-verification programs into self-verification and self-refinement, achieving effective self-correction.

\textbf{Integration of LLM with Programs.}
Several studies have enhanced LLMs by introducing programs or symbolic solvers. Some works integrate code executor or symbolic solving tools within the LLM's forward reasoning to address mathematical or symbolic reasoning problems, such as PAL~\citep{pmlr-v202-gao23f}, POT~\citep{chen2023program}, and ToRA~\citep{gou2024tora}. Others~\citep{chen2024teaching,dong2024toward,qiao-etal-2024-making,dong2024selfplayexecutionfeedbackimproving} use programs to assist in data synthesis or training; for example, AutoIf~\citep{dong2024selfplayexecutionfeedbackimproving} filters synthetic training data through testing programs. Additionally, \citet{lyu2024largelanguagemodelscode} explore the capability of LLMs to act as code interpreters to execute LeetCode programs.
Distinguishing these studies, our work is unique in that it: (1) employs verification programs for the self-correction phase; (2) utilizes pseudo-programs that do not require strict executability; (3) enables LLMs to self-generate and self-execute programs without the necessity for actual symbolic tools.

\textbf{Inference Framework.}
Unlike directly generating answers, many studies have explored enhancing LLMs' ability to solves complex tasks through reasoning frameworks. Decomposition-based methods like COT, TOT, and POT~\citep{wei2022chain,yao2024tree,chen2023program,qiao2024we} guide models to break down tasks step-by-step, while sampling-based methods~\citep{wang2022self,zhang2024accessing,dong2024progressive} explore diverse reasoning paths and select the most consistent or optimal response. Another key approach is the mechanism of reflection and self-correction~\citep{zhang-etal-2024-self-contrast,shinn2024reflexion,pan-etal-2024-automatically}, which encourages models to iteratively evaluate and refine their responses. Additionally, to balance efficiency, some works~\citep{yao2024hdflow} dynamically combine fast and slow reasoning based on task complexity. ProgCo can be seen as a method of fast and slow reasoning, where RrogVe filters out complex tasks, and ProgRe uses dual optimization for slow reasoning. Moreover, since ProgCo focuses on the self-correction, it is orthogonal to forward reasoning methods like sampling and decomposition, and can be well combined with them.

\section{Details of Method}
\label{app:method_details}

\subsection{Details of Prompts}
\label{subapp:method_prompt}

We present the prompts used for mathematical tasks in Figures \ref{fig:gen_code_prompt}, \ref{fig:execute_code_prompt}, \ref{fig:get_feedback}, \ref{fig:response_reflex_prompt}, \ref{fig:contrast_prompt}, \ref{fig:code_reflex_prompt}.

\subsection{Pseudo-code}
\label{subapp:method_pesudo_code}
We summarize the pseudo-code of ProgCo in Algorithm \ref{alg:ProgCo}.

\begin{algorithm*}
\caption{: Program-driven Self-Correction (ProgCo)}
\label{alg:ProgCo}
\begin{algorithmic}[1]
\Require {input $x$, model $M$, prompts $\{P^{gen}_{fuc},P^{exec}_{fuc},P_{fb},P_{reflex},P_{cont},P^{code}_{reflex}\}$}
\Ensure  final response $y_{final}$
\State $y_0=M(x)$ \Comment{Initial generation}
\For {iteration $i \in 0,1,...,I$}
\State {$f_{i}=M(P^{gen}_{fuc}||x)$ \Comment{Verification Program Generation (Eqn.1)}}
\State {$r_{i}=M(P^{exec}_{fuc}||x||y_i)$ \Comment{Verification Program Execution (Eqn.2)}}
\State {$fb_{i}=M(P_{fb}||x||y_i||r_i)$}
    \If {$fb_{i} = \text{``pass''}$}
        \State break
    \Else
        \State $y^{temp}_{i+1}=M(P_{reflex}||x||y_i||fb_i)$ \Comment{Preliminary Reflection (Eqn.3)}
        \If {$y^{temp}_{i+1}$ != $y_{i}$}
            \State $ins=M(P_{cont}||y_{i}||y^{temp}_{i+1})$ \Comment{Contrast and Regenerate (Eqn.4)}
            \State $y_{i+1}=M(ins||x)$
        \Else
            \State $y_{i+1}=y^{temp}_{i+1}$
        \EndIf
        \State $f_{i+1}=M(P^{code}_{reflex}||x||f_{i}||y_{i}||fb_{i})$ \Comment{Verification Program Refinement (Eqn.5)}
    \EndIf
\EndFor
\State $y_{final}=y_{i}$
\end{algorithmic}
\end{algorithm*}

\section{Details of Experiment}
\label{app:experiment_details}

\subsection{Details of Datasets}
\label{subapp:dataset_details}
The introduction to the evaluation datasets is as follows:
\begin{itemize}
\item \textbf{IFEval}~\citep{zhou2023instructionfollowingevaluationlargelanguage}: IFEval is one of the most commonly used instruction-following benchmark for LLMs, containing over 500 test samples and covering 25 types of atomic instructions.
The Prompt(Strict) metric calculates the proportion of strict following of the input prompt across all samples. For a sample, the prompt is considered followed only if all atomic instructions in it are followed. The Instruction(Strict) metric calculates the proportion of strict following of each atomic instruction across all samples.
\item \textbf{GSM8K}~\citep{cobbe2021trainingverifierssolvemath}:
GSM8K is a classic mathematical reasoning evaluation dataset that primarily focuses on math problem-solving at the grade school level. The dataset contains over 8,000 samples, with 7,473 samples in the training set and 1,319 samples in the test set.
\item \textbf{MATH}~\citep{hendrycksmath2021}:
The MATH dataset consists of 12,500 complex math competition problems covering various branches of mathematics, including algebra, geometry, probability, and number theory. The MATH training set contains 7,500 samples, and the test set includes 5,000 samples. We randomly select 500 from the test set for evaluation.
\end{itemize}

\subsection{Details of Baselines}
\label{subapp:baseline_details}
The introduction to the baselines is as follows:
\begin{itemize}
    \item \textbf{Vanilla-Reflex}: Vanilla-Reflex is a simple setup that iteratively prompts LLMs to reflect on its original output and generate a new output without an early stopping mechanism.
    \item \textbf{Self-Refine}~\citep{madaan2024self}: Self-refine iteratively examines its own output to obtain feedback and suggests refining its output based on this feedback until the examination indicates that there are no errors.
    \item \textbf{Self-Reflection}~\citep{shinn2024reflexion}: For erroneous outputs from environmental feedback, Reflection will first reflect on the output and generate task experiences. These experiences will be incorporated into the next attempt as a way of refinement. In the absence of environmental feedback, we use COT checks consistent with Self-Refine.
    \item \textbf{CheckList}~\citep{cook2024tickingboxesgeneratedchecklists}: CheckList first generates a verification list for the input task and then verifies whether the response satisfies all the verification points one by one. Any unsatisfied verification points will be used as feedback to optimize its own output.
\end{itemize}

\subsection{Details of Implementation}
\label{subapp:implementation_details}
We use azure GPT-3.5-Turbo-0613 (16K) for GPT-3.5 and GPT-4o-0806 for GPT-4o~\cite{GPT4o}. For Llama3.1-8B-Instruct~\citep{llama31}, we employ FastTransformer\footnote{https://github.com/NVIDIA/FasterTransformer} to accelerate inference. The inference temperature is set to 0, while other parameters are kept at their default values. We utilize few-shot demonstrations for both the baselines and ProgCo, adjusting the number of demonstrations to 1-3 based on the demonstration content length.

For instruction following, we use the official IFEval~\citep{zhou2023instructionfollowingevaluationlargelanguage} evaluation with Strict Prompt and Strict Instruction metrics. To achieve more accurate mathematical evaluation, we prompt GPT-4o-mini~\cite{GPT4o} to extract LLM's output answers and perform regex matching with the standard answers. Answers deemed incorrect by regex are further evaluated by GPT-4o-mini to determine their correctness, preventing errors in synonymous expressions, such as decimals and fractions.

\clearpage

\lstset{
  breaklines=true, 
  basicstyle=\ttfamily\scriptsize, 
  breakindent=0pt,
  escapeinside={(*@}{@*)}, 
}

\begin{figure*}
\begin{tcolorbox}[
colback=white!10!white,
colframe=black!75!black,
left=0.5pt, right=0.5pt,
boxsep=1pt,
title=Generate Verification Program,
breakable]
\vspace{-0.3cm}
\begin{lstlisting}
You are an expert in reverse reasoning verification. Given a problem, you need to generate a reverse verification executable Python function for that problem.

[Reverse Reasoning Introduction]
* Reverse reasoning is a method of thinking that starts from the result and verifies the problem backwards. Specifically, it involves:

1. Starting with the given answer, rather than the initial conditions of the problem.
2. Assuming this answer is correct, then conducting reverse deduction based on this assumption and the known conditions in the problem.
3. Through this reverse deduction, checking whether other known conditions or constraints in the problem can be satisfied.
4. If the results of the reverse deduction can satisfy all conditions, then the original answer can be considered correct.

For example, using the answer and known conditions 1 and 2 as assumptions, reverse reason to check if it satisfies known condition 3 in the problem statement.

To illustrate, consider a problem of solving a quadratic equation ax^2+bx+c=0:
- Forward thinking would start with a, b, c values and use the quadratic formula to calculate x.
- Reverse reasoning would:
1. Start with a possible solution x
2. Substitute it into ax^2 + bx + c
3. Check if the result equals 0
4. If it equals 0, the solution is verified as cor

[Requirements]
1. The verification function should start with the input answer and use reverse reasoning to validate the correctness of the answer.
2. The verification function should only accept one input (the answer) and output the verification result as True/False.
3. After generating the verification function name, please first write the reverse analysis verification approach as code comments, then generate the content of the verification function. Please do not output any other content.
\end{lstlisting}
\vspace{-0.3cm}
\end{tcolorbox}
\vspace{-0.5cm}
\caption{The prompt for generating verification programs.}
\label{fig:gen_code_prompt}
\end{figure*}

\begin{figure*}
\begin{tcolorbox}[
colback=white!10!white,
colframe=black!75!black,
left=0.5pt, right=0.5pt,
boxsep=1pt,
title=Execute Verification Program,
breakable]
\vspace{-0.3cm}
\begin{lstlisting}
You are a verification expert proficient in code execution and possessing extensive world knowledge, and you have the following advantages:
1. Ability to execute code flexibly, unrestricted by strict syntax or standards
2. Capacity to comprehend code purpose, overall logic, and potential issues through analysis of comments and context

Your task is to verify a given mathematical problem and its solution. A data annotator has written verification code for this problem. Please proceed as follows:

1. Carefully read the provided problem, solution process, and answer
2. Execute the verification code step by step
3. If you discover any issues in the verification code, make appropriate revisions before continuing execution
4. Derive the final verification result

Input information:
[Problem]
{query}
[Solution Process]
{response}
[Solution Answer]
{result}
[Verification Code]
{validate_response_fuc}

Please output your analysis and results strictly in the following format, without adding any additional content:

[Execution of Verification Code]
{Detailed step-by-step execution process}
[Verification Result]
{True or False}
\end{lstlisting}
\vspace{-0.3cm}
\end{tcolorbox}
\vspace{-0.5cm}
\caption{The prompt for executing verification programs.}
\label{fig:execute_code_prompt}
\end{figure*}

\begin{figure*}
\begin{tcolorbox}[
colback=white!10!white,
colframe=black!75!black,
left=0.5pt, right=0.5pt,
boxsep=1pt,
title=Get Feedback,
breakable]
\vspace{-0.3cm}
\begin{lstlisting}
Extract key information from the execution process of the verification code and convert it into natural language form.
[Problem]
{query}
[Execution of Verification Code]
{execute_content}
\end{lstlisting}
\vspace{-0.3cm}
\end{tcolorbox}
\vspace{-0.5cm}
\caption{The prompt used to get feedback from execution.}
\label{fig:get_feedback}
\end{figure*}

\begin{figure*}
\begin{tcolorbox}[
colback=white!10!white,
colframe=black!75!black,
left=0.5pt, right=0.5pt,
boxsep=1pt,
title=Preliminary Reflection of Response,
breakable]
\vspace{-0.3cm}
\begin{lstlisting}
Your task is to reflex whether a solution is correct.

Given a problem [Problem], a reverse reasoning validation expert generated an executable Python function for reverse validation [Initial Verification Code] of this problem. 
However, the solution has not been validated by the verification function. This means that either the solution or the verification function, or both, contain errors.

You need to carefully analyze and complete the following tasks:
1. Reflect on the initial solution process:
   - Compare in detail the solution and the feedback, examine the logic and accuracy of the solution approach **step by step**, consider whether the errors lie with the feedback or with the solution.
   - If errors or inadequacies are found, provide detailed feedback and suggestions for improvement
   - If it is found that there are no errors in the solution but rather errors exist in the feedback, state that the solution is correct.
2. Provide a new solution:
   - If the initial solution is correct, please use the original solution process (Note: If the verification code is correct, you also need to repeat the initial verification code word for word)
   - If errors or inadequacies are found, revise the solution based on your reflection.

Note: 
- You cannot refuse to generate a new solution due to missing information or other reasons.
Please strictly output your analysis and revision according to the following format, without any additional content:

[Reflection]
{Detailed reflection process}
[New Solution]
{Complete solution process based on the reflection}
\end{lstlisting}
\vspace{-0.3cm}
\end{tcolorbox}
\vspace{-0.5cm}
\caption{The prompt for reflecting on responses.}
\label{fig:response_reflex_prompt}
\end{figure*}

\begin{figure*}
\begin{tcolorbox}[
colback=white!10!white,
colframe=black!75!black,
left=0.5pt, right=0.5pt,
boxsep=1pt,
title=Contrast on pre and post-reflection responses,
breakable]
\vspace{-0.3cm}
\begin{lstlisting}
You are an expert at comparing and extracting key points.
Task: Analyze the differences between two solutions and extract key points
Background: For the same problem, two solutions have provided different answers. We need to analyze these differences in depth to identify the key aspects of the problem.

Steps:
1. Carefully read the problem and both solutions
2. Ignore surface differences in expression, focus on substantial differences in content and method
3. Compare the core ideas, key steps, and final results of both solutions
4. Summarize the essence of the problem reflected by these differences and the key points to note when solving

Output requirements:
1. Concisely list 1-3 key points
2. Each point should be specific and helpful for regenerating a better solution

Output format:
[Comparative Analysis Process]
{Your comparative analysis process}
[Core Differences in Solutions]
{Summarize the differences in solutions based on the comparative analysis process, answer in bullet points}
[Key Points to Note When Solving the Problem]
{Summarize the key points to note when solving the problem based on the differences in solutions, answer in bullet points}
\end{lstlisting}
\vspace{-0.3cm}
\end{tcolorbox}
\vspace{-0.5cm}
\caption{The prompt for contrasting responses and generating insights.}
\label{fig:contrast_prompt}
\end{figure*}

\begin{figure*}
\begin{tcolorbox}[
colback=white!10!white,
colframe=black!75!black,
left=0.5pt, right=0.5pt,
boxsep=1pt,
title=Reflection on the verification program,
breakable]
\vspace{-0.3cm}
\begin{lstlisting}
Your task is to reflex whether a verification code is correct.

[Reverse Reasoning Introduction]
* Reverse reasoning is a method of thinking that starts from the result and verifies the problem backwards. Specifically, it involves:

1. Starting with the given answer, rather than the initial conditions of the problem.
2. Assuming this answer is correct, then conducting reverse deduction based on this assumption and the known conditions in the problem.
3. Through this reverse deduction, checking whether other known conditions or constraints in the problem can be satisfied.
4. If the results of the reverse deduction can satisfy all conditions, then the original answer can be considered correct.

For example, using the answer and known conditions 1 and 2 as assumptions, reverse reason to check if it satisfies known condition 3 in the problem statement.

To illustrate, consider a problem of solving a quadratic equation ax^2 + bx + c = 0:
- Forward thinking would start with a, b, c values and use the quadratic formula to calculate x.
- Reverse reasoning would:
1. Start with a possible solution x
2. Substitute it into ax^2 + bx + c
3. Check if the result equals 0
4. If it equals 0, the solution is verified as cor

Given a problem [Problem], a reverse reasoning validation expert generated an executable Python function for reverse validation [Initial Verification Code] of this problem. However, the verification code did not pass when verifying a solution [Solution] of the problem.

The error may come from the solution or from the verification code.
You need to carefully analyze and complete the following tasks:
1. Reflect on the initial verification code:
   - First, examine the logic and accuracy of the verification code step by step
   - If errors or inadequacies are found, provide detailed feedback and suggestions for improvement
2. Provide a new verification code:
   - If the initial verification code is correct, you can use the original verification code
   - If errors or inadequacies are found, revise the verification code based on your reflection
   - Keep detailed comments in the revised validation code

Please strictly output your analysis and revision according to the following format, without any additional content:

[Reflection]
{Detailed reflection process}
[New Verification Code]
{Complete verification code based on the reflection}
\end{lstlisting}
\vspace{-0.3cm}
\end{tcolorbox}
\vspace{-0.5cm}
\caption{The prompt for reflecting on verification programs.}
\label{fig:code_reflex_prompt}
\end{figure*}

\begin{figure*}
\begin{tcolorbox}[
colback=white!10!white,
colframe=black!75!black,
left=0.5pt, right=0.5pt,
boxsep=1pt,
title=ProgVe Case on IFEval (GPT-3.5),
breakable]
\vspace{-0.3cm}
\begin{lstlisting}
(*@\textbf{Instruction:}@*) Write a joke about a startup that sells dog food in a song. Your entire response should be in English, and in all capital letters. Your answer must contain a title, wrapped in double angular brackets, i.e. <<title>>.

(*@\textbf{Response:}@*)
<<title>> "Barking Business"
Why did the startup that sells dog food decide to sing about their business?
Because they wanted to unleash their catchy jingle and make tails wag to the beat!

(*@\textbf{Verification Program:}@*)
def validate_response(response):
    errors = []
    # Check if the response is in all capital letters
    if response != response.upper():
        errors.append("Response is not entirely in capital letters")
    # Check if the response is in English
    if not is_english(response):
        errors.append("Response is not in English")
    # Check if the response contains a title wrapped in double angular brackets
    if not has_title(response):
        errors.append("Response does not contain a title wrapped in double angular brackets")
    # If errors exist, return False and error messages; otherwise return True
    if errors:
        return False, "\n".join(errors)
    return True, "No Error"

# Auxiliary functions:
# is_english(text): Check if the text is in English
# has_title(text): Check if the text contains a title wrapped in double angular brackets

(*@\textbf{Verification Program Execution:}@*)
Step 1: Initialize errors list
errors = []
Step 2: Check if the response is in all capital letters
If response != response.upper():
    errors.append("Response is not entirely in capital letters")
The response is not entirely in capital letters,so this error is added to the list.
Step 3: Check if the response is in English
Assuming is_english(response) returns True (as it appears to be in English), no error is added.
Step 4: Check if the response contains a title wrapped in double angular brackets
If not has_title(response):
    errors.append("Response does not contain a title wrapped in double angular brackets")
The response does not contain a title wrapped in double angular brackets, so this error is added to the list.
Step 5: Return result
errors list contains two items: "Response is not entirely in capital letters" and "Response does not contain a title wrapped in double angular brackets"
Since errors is not empty, the function returns False and the error messages.

So, the final result of validate_response(response) is:
<result>False, "Response is not entirely in capital letters\nResponse does not contain a title wrapped in double angular brackets"</result>

(*@\textbf{Feedback:}@*)
The initial response needs to be modified based on the following suggestions:
1. Ensure that the response is entirely in capital letters.
2. Include a title wrapped in double angular brackets.

\end{lstlisting}
\vspace{-0.3cm}
\end{tcolorbox}
\vspace{-0.5cm}
\caption{A case of ProgVe self-verifying on IFEval.}
\label{fig:ifeval_case}
\end{figure*}

\begin{figure*}
\begin{tcolorbox}[
colback=white!10!white,
colframe=black!75!black,
left=0.5pt, right=0.5pt,
boxsep=1pt,
title=ProgVe Case on GSM8K (GPT-3.5),
breakable]
\vspace{-0.3cm}
\begin{lstlisting}
(*@\textbf{Question:}@*)
Marissa is hiking a 12-mile trail. She took 1 hour to walk the first 4 miles, then another hour to walk the next two miles. If she wants her average speed to be 4 miles per hour, what speed (in miles per hour) does she need to walk the remaining distance?

(*@\textbf{Response:}@*) 
Marissa has already walked 4 + 2 = <<4+2=6>>6 miles.
She has 12 - 6 = <<12-6=6>>6 miles left to walk.
She has spent 1 + 1 = <<1+1=2>>2 hours walking so far.
To average 4 miles per hour, she needs to walk the remaining 6 miles in 6 / 4 = <<6/4=1.5>>1.5 hours.
Thus, she needs to walk the remaining distance at a speed of 6 / 1.5 = <<6/1.5=4>>4 miles per hour. Answer: \boxed{4}.

(*@\textbf{Verification Program:}@*)
def verify_remaining_speed(remaining_speed):
    # Known conditions
    total_distance = 12
    first_distance = 4
    first_time = 1
    second_distance = 2
    second_time = 1
    target_average_speed = 4

    # Calculate the time taken to walk the remaining distance
    remaining_time = (total_distance - first_distance - second_distance) / remaining_speed

    # Calculate the total time taken to walk the entire trail
    total_time = first_time + second_time + remaining_time

    # Calculate the average speed based on the total time and total distance
    average_speed = total_distance / total_time

    # Check if the average speed is 4 miles per hour
    if abs(average_speed - target_average_speed) > 0.01:  # Allow for small rounding errors
        return False

    # All conditions are satisfied
    return True

(*@\textbf{Execution of Verification Program:}@*)
Calculate the time taken to walk the remaining distance:
remaining_time = (total_distance - first_distance - second_distance) / remaining_speed
remaining_time = (12 - 4 - 2) / 4
remaining_time = 6 / 4
remaining_time = 1.5

Calculate the total time taken to walk the entire trail:
total_time = first_time + second_time + remaining_time
total_time = 1 + 1 + 1.5
total_time = 3.5

Calculate the average speed based on the total time and total distance:
average_speed = total_distance / total_time
average_speed = 12 / 3.5
average_speed = 3.4285714285714284

Check if the average speed is 4 miles per hour:
abs(average_speed - target_average_speed) = abs(3.4285714285714284 - 4) = 0.5714285714285716
0.5714285714285716 > 0.01, so this check fails.

The condition in step 5 fails, therefore the function returns False.

(*@\textbf{Feedback:}@*)
We know that Marissa is hiking a 12-mile trail. She took 1 hour to walk the first 4 miles and another hour to walk the next 2 miles. Her target average speed is 4 miles per hour.
First, let's calculate the time taken to walk the remaining distance. The remaining distance is 12 miles - 4 miles - 2 miles = 6 miles. If Marissa wants her average speed to be 4 miles per hour, the time taken to walk the remaining distance would be 6 miles / 4 miles per hour = 1.5 hours.
Next, let's calculate the total time taken to walk the entire trail. The total time is the sum of the time taken for the first part, the second part, and the remaining part. So, the total time is 1 hour + 1 hour + 1.5 hours = 3.5 hours.
Then, let's calculate the average speed based on the total time and total distance. The average speed is the total distance divided by the total time. So, the average speed is 12 miles / 3.5 hours = 3.4285714285714284 miles per hour.
However, the target average speed is 4 miles per hour, which does not match the calculated average speed of 3.4285714285714284 miles per hour. Therefore, the solution of 4 miles per hour for the remaining speed fails our verification process because it leads to inconsistencies with the known facts.
\end{lstlisting}
\vspace{-0.3cm}
\end{tcolorbox}
\vspace{-0.5cm}
\caption{A case of ProgVe self-verifying on GSM8K.}
\label{fig:gsm8k_case_verification}
\end{figure*}

\begin{figure*}
\begin{tcolorbox}[
colback=white!10!white,
colframe=black!75!black,
left=0.5pt, right=0.5pt,
boxsep=1pt,
title=ProgRe Case on GSM8K following ProgVe (GPT-3.5),
breakable]
\vspace{-0.3cm}
\begin{lstlisting}
(*@\textbf{Preliminary Reflection of Response:}@*)
[Reflection]
The initial solution correctly calculates the remaining distance to be 6 miles (12 miles - 4 miles - 2 miles) and determines that Marissa needs to walk the remaining 6 miles in 1.5 hours to achieve an average speed of 4 miles per hour. However, the feedback points out that the calculated average speed of 3.4285714285714284 miles per hour does not match the target average speed of 4 miles per hour.
Upon reviewing the feedback, it is clear that the initial solution made an error in assuming that Marissa needs to walk the remaining distance at a speed of 4 miles per hour. This assumption is incorrect because the target average speed of 4 miles per hour refers to the entire trail, not just the remaining distance.
[New Solution]
To find the speed Marissa needs to walk the remaining distance, we can calculate the time it took her to walk the first 6 miles and subtract it from the total time.
Marissa took 1 hour to walk the first 4 miles and another hour to walk the next 2 miles, so she took a total of 1 + 1 = 2 hours to walk the first 6 miles.
The remaining distance is 12 miles - 6 miles = 6 miles.
To find the speed, we divide the remaining distance by the remaining time:
Speed = Remaining Distance / Remaining Time
Speed = 6 miles / 2 hours
Speed = 3 miles per hour
Therefore, Marissa needs to walk the remaining distance at a speed of 3 miles per hour. Answer: \boxed{3}.

(*@\textbf{Contrast Insights:}@*)
- Ensure accurate calculation of the remaining time by considering the total time spent and the desired average speed.
- Double-check calculations to ensure that the remaining speed is correctly calculated by dividing the remaining distance by the remaining time.

(*@\textbf{New Response:}@*)
To find the speed Marissa needs to walk the remaining distance, we first need to calculate the remaining time. 
Marissa has already walked 4 miles in 1 hour and another 2 miles in another hour, so she has already spent 1 + 1 = 2 hours on the trail. 
Since she wants her average speed to be 4 miles per hour, we can calculate the total time she should spend on the trail by dividing the total distance (12 miles) by the desired average speed (4 miles per hour). 
Total time = Total distance / Average speed
Total time = 12 miles / 4 miles per hour
Total time = 3 hours
Since Marissa has already spent 2 hours on the trail, the remaining time is 3 hours - 2 hours = 1 hour. 
To find the speed Marissa needs to walk the remaining distance, we divide the remaining distance (12 miles - 4 miles - 2 miles = 6 miles) by the remaining time (1 hour). 
Remaining speed = Remaining distance / Remaining time
Remaining speed = 6 miles / 1 hour
Remaining speed = 6 miles per hour
Therefore, Marissa needs to walk the remaining distance at a speed of 6 miles per hour.

(*@\textbf{Verification Program Refinement:}@*)
[Reflection]
The initial verification code attempts to validate the solution by following a reverse reasoning approach. However, there is an error in the logic used to check if the average speed is 4 miles per hour.
Here is a detailed reflection of the code:
- The code properly calculates the time taken to walk the remaining distance as `(total_distance - first_distance - second_distance) / remaining_speed`.
- It then calculates the total time taken to walk the entire trail as the sum of the first time, second time, and remaining time.
- The code calculates the average speed based on the total time and total distance using the formula `average_speed = total_distance / total_time`.
- However, the check for the average speed being 4 miles per hour is incorrect. It compares the calculated average speed with the target average speed using a tolerance of `0.01`. This is incorrect as there is no need for a tolerance in this case. The two values should be exactly equal for validation.
[New Verification Program]:
def verify_remaining_speed(remaining_speed):
    # Known conditions
    total_distance = 12
    first_distance = 4
    first_time = 1
    second_distance = 2
    second_time = 1

    # Calculate the time taken to walk the remaining distance
    remaining_time = (total_distance - first_distance - second_distance) / remaining_speed

    # Calculate the total time taken to walk the entire trail
    total_time = first_time + second_time + remaining_time

    # Calculate the average speed based on the total time and total distance
    average_speed = total_distance / total_time

    # Check if the average speed is 4 miles per hour
    if average_speed != 4:
        return False

    # All conditions are satisfied
    return True
\end{lstlisting}
\vspace{-0.3cm}
\end{tcolorbox}
\vspace{-0.5cm}
\caption{A case of ProgRe self-refining on GSM8K (following ProgVe).}
\label{fig:gsm8k_case_refine}
\end{figure*}

\end{document}